\documentclass[11pt,a4paper]{article}
\usepackage[margin=1in]{geometry}

\usepackage{algorithm,algpseudocode}
\usepackage{amsmath,amsfonts,amssymb}
\usepackage{bm}
\usepackage{comment}
\usepackage{float}
\usepackage{graphicx}
\usepackage{hyperref}
\usepackage{latexsym}
\usepackage{makecell}
\usepackage[section]{placeins}
\usepackage{subcaption}
\usepackage{tabularx}
\usepackage{times}
\usepackage{xcolor}
\usepackage{url}

\usepackage{natbib}

\usepackage{pgf}
\usepackage{tikz}
\usetikzlibrary{positioning}

\usepackage[bottom]{footmisc}
\makeatletter
\def\blfootnote{\xdef\@thefnmark{}\@footnotetext}
\makeatother

\title{Semantic Matching Against a Corpus: New Applications and Methods}

\author{Lucy H.~Lin$^\star$ \quad Scott Miles$^\dagger$ \quad Noah A.~Smith$^{\star, \diamond}$\\[5pt]
  $^\star$Paul G.~Allen School of Computer Science \& Engineering, University of Washington\\
  $^\dagger$Human Centered Design \& Engineering, University of Washington\\
  $^\diamond$Allen Institute for Artificial Intelligence\\[5pt]
  {\tt \{lucylin,nasmith\}@cs.washington.edu}, \\ {\tt milessb@uw.edu} \\}

\date{}

\begin{document}
\maketitle

\begin{abstract}
We consider the case of a domain expert who wishes to explore the extent to which a particular idea is expressed
in a text collection.  We propose the task of semantically matching the idea, expressed as a natural language proposition, against a corpus.  We create two preliminary tasks derived from existing datasets, and then introduce a more realistic one on disaster recovery designed for
emergency managers, whom we engaged in a user study.  On the latter, we find that a new model built from natural language entailment data produces higher-quality matches than simple word-vector averaging, both on expert-crafted queries and on ones produced by the subjects themselves. This work provides a proof-of-concept for such applications of semantic matching and illustrates key challenges.

\blfootnote{Code and data for this paper is provided at: \\
\url{https://homes.cs.washington.edu/~lucylin/research/semantic\_matching.html}}

\end{abstract}

% *****************************************************************

\section{Introduction}
\label{sec:introduction}
Extensive recent work contributes computational models of sentence-level
entailment, and proposition-level semantic similarity more broadly.
We propose that end-users of these methods could eventually include:
(i) historians of science tracking
expression of the idea that ``vaccines cause autism'' after the
1998 study in \emph{The Lancet} making this claim;
(ii) political scientists and journalists tracking fine-grained opinions like
``immigrants are often unfairly used as scapegoats for problems in
society'' in the media;
and (iii) public servants seeking to understand the challenges facing a
community after a disaster by tracking claims like ``dealing with authorities
is causing stress and anxiety.''

What all of these examples have in common is that a user specifies a natural
language \textbf{proposition query}: an idea likely to to occur in a given
text collection.
Natural languages offer many ways to express any idea, so it is an open
question what kinds of semantic matching methods will be required to fulfill
the information needs of different kinds of users.

\begin{figure}[t]
\centering
\begin{tikzpicture}
\tikzstyle{every node}=[font=\small]

\node[rectangle,draw=black,align=left,text width=0.75\textwidth] (prototype)
    {\textbf{Proposition query ($s_p$):} \\
    ``Dealing with authorities is causing stress and anxiety.''};

\node[rectangle,align=left,draw=black,text width=0.75\textwidth] (sentences) [below=.6in of prototype]
    {\textbf{Matched sentences ($C_m$):} \\
     ``Relationships Aotearoa... said it is unfamiliar bureaucratic
       systems which are causing the majority of the stress.''\\[0.5em]
     ``... those in charge of the earthquake recovery are making
       moves to appease the growing anger among homeowners.'' \\
   };

\begin{scope}[transform canvas={xshift=-.75in}]
\path[->,line width=.5mm] (prototype) edge node[right,align=left] {query corpus ($C$) \\via matching function $m$} (sentences);
\end{scope}
\end{tikzpicture}

\caption{An example of semantic matching in the domain of natural disaster recovery.
\label{fig:introex}}
\end{figure}
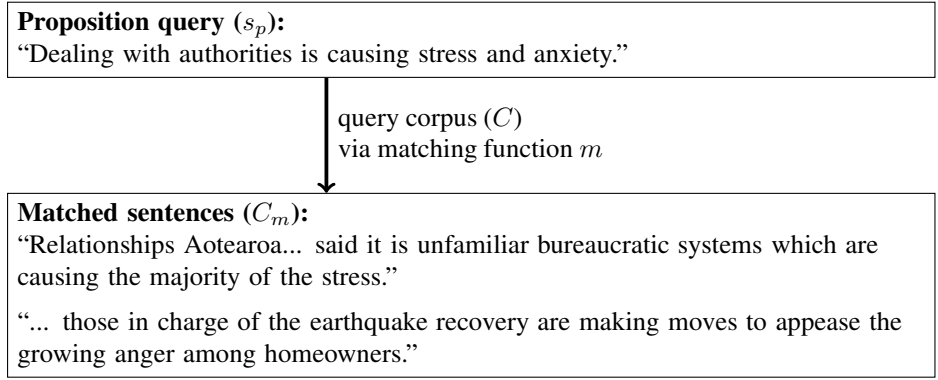

In this paper, we demonstrate how semantic matching methods can be used more
widely.
In \S\ref{sec:simple}, we start with simple word-vector-based methods inspired
by earlier work on semantic similarity and relatedness.
As an initial test, we define two new tasks that
exploit existing labeled corpora (\S\ref{subsec:simple-existing}).
Using the CNN/Daily Mail Reading Comprehension dataset
\citep{Hermann:15}, we evaluate whether a model can identify the relevant
sentence in an article given a proposition query.
Second, using the Media Frames Corpus (a collection of news articles about
immigration in the US; \citealp{Card:15}),
we derive proposition queries from the examples in
its annotation codebook.  In both evaluations, we find that a simple
word vector average-based matching algorithm
can retrieve sentences marked by
annotators reasonably well.\footnote{Note that these are not the
  conventional tasks associated with these two corpora.}

Given those positive results, we introduce a more realistic application:
propositions in the domain of natural disaster recovery
(\S\ref{subsec:simple-expert}; example in Fig.~\ref{fig:introex}).
A domain expert collaborating on the research provided the proposition queries,
and our evaluation is a user study with twenty emergency managers.

Since this application suggests that a more nuanced model of semantics than the
word-vector-averaging models is necessary, we turn to more complex
entailment-based models (\S\ref{sec:complex}).
We introduce a new syntax-based model for matching,
trained on the SNLI dataset \citep{Bowman:15}.
Our user study shows that this model offers higher quality matches than the vector-averaging baselines.
We find further confirmation of these results in a follow-up study
where the emergency managers themselves created the proposition queries.
Finally, we introduce an application of semantic matching, \emph{semantic
measurement} (\S\ref{sec:measurement}), by qualitatively exploring the
frequency of matches over time.
In the user community we surveyed, we find that there is
interest in tools for semantic matching and measurement.

\section{Problem Formulation}
\label{sec:problem}
We formalize the semantic matching problem as follows. Let $C$ denote a corpus
consisting of a collection of documents, each a list of English sentences
(individually denoted by $s$). $s_p$ will be the proposition query, also a
sentence.

The goal is to find sentences $s \in C$ such that $s$ expresses the
idea contained in $s_p$.  To do so, we assume that sentences $s \in C$ will be
ranked by some function $m(s_p, s)$ and the top $n$ will be returned to the
user (as the set $C_m$).  We can think of $m$ as a model of semantic
similarity (as in \S\ref{sec:simple}) or entailment (as in
\S\ref{sec:complex}). This setup is quite similar to (sentence-level) text retrieval,
except that the user is assumed to be interested in the full set $C_m$, rather
than answering a specific information need using \emph{any} relevant match.
(See \S\ref{sec:related} for further discussion of related tasks.)

We note that our approach assumes segmentation at the sentence level,
but alternative formulations (where the expression of an idea may span
several sentences or only a clause or phrase in a sentence) can be
considered straightforwardly.  Here, document structure is not used in
identifying matches, but could be an interesting source of information in
future work.

\section{Preliminary Models \& Experiments}
\label{sec:simple}

We begin with simple word-vector-averaging models.
We then construct two relevant tasks based on existing corpora (CNN/Daily Mail
Reading Comprehension, Media Frames Corpus), and demonstrate the models'
viability on these tasks.
Given these results, we introduce a more realistic application, where
a domain expert specifies proposition queries about natural disaster recovery,
and validate the output by performing a user study with emergency response
professionals.

\subsection{Word Vector Averaging}
\label{subsec:simple-model}

To match proposition queries $s_p$ to sentences $s$ from a corpus, we first consider a scoring method inspired by work on paraphrase \citep{Wieting:16} and averaging networks \citep{Iyyer:15}.  Each sentence is represented as the average of its word vectors, and the similarity score between $s_p$ and $s$ is the cosine similarity between their vectors.

Of course, the choice of pre-trained word vectors could have a large effect on the quality of a semantic matching system, so we examine two options.
We first consider 300-dimensional paraphrastic word vectors generated by \citet{Wieting:16}; we selected these because they were designed specifically for semantic similarity between sequences.
We also select the widely used word2vec vectors \citep{Mikolov:13}, which are trained on Google News and contain 300-dimensional vectors for approximately 3 million words.\footnote{\url{https://code.google.com/archive/p/word2vec/}} These are of interest because they are relatively fast to train on large amounts of data.  Because they are derived from unstructured news text, they are more likely to contain proper nouns/entities of interest than the paraphrastic vectors, which are trained on the Paraphrase Database \citep{Ganitkevitch:13}.

As a sanity check, we also test an even simpler information retrieval-inspired model that uses cosine similarity of tf-idf vectors of $s_p$ and $s$.

\subsection{Matching Queries in Existing Corpora}
\label{subsec:simple-existing}

Before investing in the design of a new application,
we exploit existing corpora labeled for related tasks (CNN/Daily Mail Reading
Comprehension and Media Frames Corpus) to test the effectiveness of simple
word-vector-averaging models (\S\ref{subsec:simple-model}). In both cases, our
evaluation differs from the tasks originally introduced by the dataset,
because our interest is in semantic matching applications (\S\ref{sec:problem}).

\subsubsection{CNN/Daily Mail}

The CNN/Daily Mail Reading Comprehension dataset \citep{Hermann:15} contains 93k articles from CNN and 220k articles from the Daily Mail. Each instance consists of an article, a query (constructed from bullet point summaries in the original articles), and an answer to the query.
For each instance, we take the proposition query $s_p$ to be its query and the ``corpus'' $C$ to be the set of sentences in its article; the model is asked to find the sentence which contains the entity in the answer.\footnote{The desired entity may appear in more than one sentence, but in general only is present in a small fraction of the total sentences in an article.}
This problem is simpler than the application described in \S\ref{sec:problem}:  $s_p$ is only being matched against sentences in one document (average 30 sentences).  Nonetheless, this dataset provides an initial testbed.\footnote{\citet{Chen:16} found that in many of the ``answerable'' cases in their analysis of the CNN/Daily Mail dataset, identifying the most relevant single sentence goes a long way. While this property may work against advancing reading comprehension models, it is ideal for our evaluation.}
 We emphasize that we are interested only in identifying relevant sentences, and not in finding the answer-entity.
We consider a sentence relevant if it contains the correct answer.

\subsubsection{Media Frames Corpus}

The Media Frames Corpus \citep{Card:15} contains several thousand news articles related to three policy issues (immigration, tobacco, and same-sex marriage).  These articles were annotated with fifteen ``framing dimensions'' according to a codebook developed
by \citet{Boydstun:14}.\footnote{Readers interested in the details of the framing dimensions are referred to those works; examples of framing dimensions salient in the immigration data include \emph{fairness and equality}, \emph{crime and punishment}, and \emph{cultural identity}.}
The texts were annotated by a team of political science experts according to the framing dimensions; any span of text could be labeled with any frame, and overlapping is possible. An example span of text annotated with the \emph{quality of life} frame is ``we hear statistics rather than stories, stories of lives mired in human suffering.''

Importantly, the codebook includes expert-designed \textbf{examples} for each framing dimension.  We take proposition queries $s_p$ to be these examples.  The intuition is that a sentence in the corpus that matches a codebook example for frame $F$ is also expected to evoke frame $F$.
For instance, ``immigration rules have changed unfairly over time'' and ``allowing unauthorized immigration is unfair to those who apply and wait'' are both examples of the \emph{fairness and equality} frame.

In this work, we focus on the immigration-related articles, as the codebook for this subset of the corpus was most complete.
From the codebook, we obtain 30 proposition queries across ten framing dimensions (not every framing dimension has examples provided for immigration). The full list of codebook examples used is provided in the appendix (Table~\ref{tab:framelist}).

Because annotated spans can be any part of a sentence, we consider a sentence to be annotated with a frame if any part of it is annotated with that frame.
In cases where the corpus annotators disagree on which framing dimension is evoked, we note agreement if any of the annotators has specified the frame of interest.\footnote{As \citet{Card:15} note, some subjectivity in frame annotations is expected, as the same text can be interpreted differently depending on the reader.}
We will examine how well the output from the models described in \S\ref{subsec:simple-model} align with existing frame annotations.  We do not expect high recall on this task, since many annotations in the corpus evoke framing dimensions in ways semantically distant from the codebook's examples.

\begin{figure*}[t]

\centering
\input{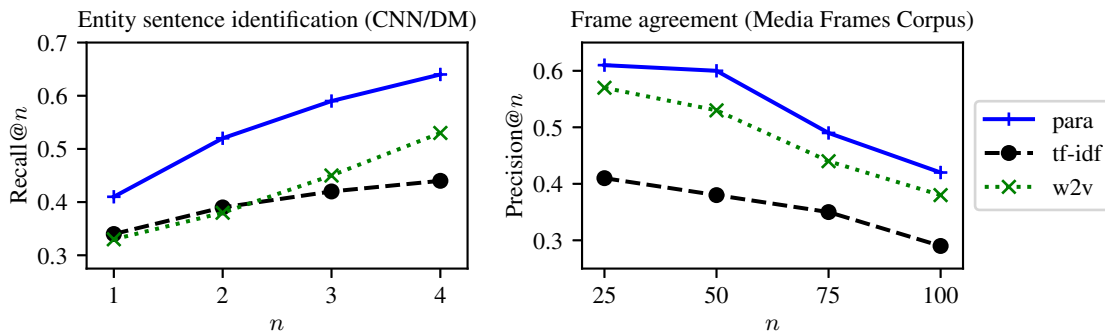}

\caption{Word-vector-averaging model results for tasks based on
  existing corpora (\S\ref{subsec:simple-existing}).}
\label{fig:existing-result}
\end{figure*}

\subsubsection{Results on Existing Corpora}
We run each of the models in \S\ref{subsec:simple-model} across the train and test partitions of the CNN/Daily Mail corpus, and on the immigration section of the Media Frames Corpus.
For the CNN/Daily Mail evaluation, we compute \emph{recall} at different values of $n$ (the number of top-scoring sentences to output) to see how well our models can identify the relevant sentence(s).
In contrast, for the Media Frames Corpus, recall is not interesting since matches to frame annotations will certainly not cover all possible evocations of their frame, so we examine \emph{precision} for varying values of $n$.

We plot the results in Fig.~\ref{fig:existing-result}. In both tasks, we find that the word-vector-based variants result in improved performance over the tf-idf baseline. (In the CNN/Daily Mail task, the word2vec and tf-idf baselines behave similarly for $n=1$ and $n=2$; as $n$ increases, word2vec becomes significantly better.) We also find that the paraphrastic vector model performs better than word2vec, which may be a result of the paraphrastic vectors being trained with semantic similarity tasks in mind.
Of course, we expect that this simple method can be improved with better sentence representations and/or application-specific supervision; we nonetheless consider these results encouraging.

\subsection{Matching Expert Queries}
\label{subsec:simple-expert}

The experiments in \S\ref{subsec:simple-existing} provide a proof of concept:  proposition queries can be matched in text using word vectors.  We now turn to a design that considers real users who seek to match ideas to text in a specific domain.
 In particular, we collaborate with an expert in disaster recovery to examine how text sources (e.g., newspapers and reports from government and utility organizations) reveal how communities recover.

\subsubsection{Domain Description and Data}

Researchers and public servants are interested in understanding the challenges facing a community after a disaster. However, on-the-ground empirical studies can be expensive to conduct, especially across a multi-year recovery period and a wide variety of variables. We propose that these users might obtain additional data through semantic matching of ideas of interest in relevant text.

More specifically, we examine recovery after the Canterbury/Christchurch (New Zealand) earthquakes that took place in late 2010 and early 2011.
We collected 982 earthquake-related articles from New Zealand news
websites,\footnote{\url{http://www.stuff.co.nz} and \url{http://www.nzherald.co.nz}} spanning 2011 through 2015.
We obtained 20 proposition queries from our domain expert; the queries cover topics like community wellbeing, infrastructure, and decision making. An example query is: ``The council should have consulted residents before making decisions.'' The full list of proposition queries is provided in Table~\ref{tab:splist} in the appendix.

\subsubsection{User Study Evaluation}

To evaluate and compare the performance of the models, we conducted a user study
with twenty emergency managers.\footnote{The emergency managers were
  solicited for this study through professional connections of our domain
  expert. Their judgments
  of our output were anonymized upon survey completion; their
  responses to a set of qualitative feedback questions
  (\S\ref{subsec:complex-result}) were not. This study was IRB-approved.}
Emergency managers are state/local personnel responsible for planning, administration, operations, and logistics related to natural and manmade hazard events, and therefore might be interested in relevant ideas found in text.

\paragraph{Experimental design.}
In this experiment, we compare word2vec and paraphrastic word-vector-averaging models; unlike the tasks in \S\ref{subsec:simple-existing}, we turn to users to evaluate the quality of matches.\footnote{Here we report only these two conditions, but in fact the user study also included the models described below in \S\ref{subsec:complex-model}, because we expected (correctly) that more powerful models would improve output quality.  We offer this simplified version of the experiment first to make the study's design clear, and describe the remainder in \S\ref{subsec:complex-eval}.}
Every sentence in the news corpus was scored against each of the 20 instances of $s_p$, for each model considered.

\paragraph{User study.}
Ideally, we would have our users judge how well every candidate sentence matches every $s_p$. Since expert users are finite, we instead sampled sentences from the following categories for the word2vec and paraphrastic vector-based models: (i) \emph{top}, the 25 highest-scoring sentences output by the model; (ii) \emph{middle}, 25 sentences, sampled randomly from those in ranks 26--250 according to the model scores; (iii) \emph{bottom}, 25 sentences, sampled randomly from those ranked at 251 or lower.

We gave each user the prompt, ``Given an idea sentence, score each candidate sentence on a 1--5 scale based on how well it expresses the idea.
The preceding and following sentences for each candidate are provided for context, but please score the quality of only the bolded candidate sentence.''\footnote{For our users' ease of understanding, we used the term ``idea sentence'' when referring
to $s_p$ instead of the more technical ``proposition query.''
In the instruction sheet, we noted that an idea sentence ``expresses a
relationship between concepts,'' but did not provide a more formal definition
to avoid overly constraining the idea sentences the participants created in
\S\ref{subsec:follow-up}.}
We provided users with a sample idea sentence and candidate sentences scored
by the same domain expert who supplied the idea sentences (Table~\ref{tab:candidateex}).
We also provided score descriptions from 1 through 5 (Table~\ref{tab:scoring}).

The candidate sentences to be scored were spread among all 20 participants;
users were not made aware of which model or sentence category the output came from.
To allow calculation of inter-annotator agreement, half of the
sentences received three judgments (rather than just one).%
\footnote{Between output from the models in \S\ref{subsec:simple-model} and
    \S\ref{subsec:complex-model}, each participant scored approximately 400
    sentences.}
We computed Krippendorf's $\alpha$ for interval data to be 0.784, which
indicates reasonable agreement when users rate the same sentence
\citep{Krippendorff:12}.

\begin{table*}[h]
{\small
\begin{tabularx}{\textwidth}{|l|X|}
\hline
\textbf{Score}	& \textbf{Example candidate (bold) and provided context} \\
\hline
\hline
1	&	The data was the latest demand and supply information on the Canterbury rebuild and wider recovery, MBIE said.\\
 & \textbf{The quarterly report for Canterbury included analysis on Greater Christchurch Value of Work, Employment and Accommodation projections.} \\
 & The forecasts were based on Canterbury Earthquake Recovery Authority projections of work to be done on the residential rebuild and repairs, infrastructure and commercial work.
\\
\hline
2   & The quarterly report for Canterbury included analysis on Greater Christchurch Value of Work, Employment and Accommodation projections. \\
 & \textbf{The forecasts were based on Canterbury Earthquake Recovery Authority projections of work to be done on the residential rebuild and repairs, infrastructure and commercial work.} \\
 & - The Press \\
\hline
3	& Migrants were now filling most of the rising number of construction jobs but beneficiaries moving into work were also contributing, MBIE’s quarterly “job-matching” report said. \\
 & \textbf{The construction sector’s workload was expected to peak in the December 2016 quarter at a value of about \$1.6 billion.} \\
 & The residential rebuild would run at “elevated levels” from 2015 until 2018 but commercial work would become increasingly important.
\\
\hline
4   & Skilled vacancies rose in most industry groups with education and training reporting the biggest increase at 2.6 percent in the month, for an annual gain of 12.4 per cent. \\
 & \textbf{The need for skilled workers increased across all occupation groups, which is divided between technician and trade workers, professionals and managers.} \\
 & Technicians and trades worker vacancies were up 2.4 per cent in December for an annual gain of 3.5 per cent. \\
\hline
5	& The additions to the current workforce of 30,000 will mostly work on commercial projects or infrastructure, the Ministry of Business, Innovation and Employment (MBIE) predicts. \\
 & \textbf{Greater Christchurch’s labour supply for the rebuild was tight and was likely to remain that way for the next three years.} \\
 & Migrants were now filling most of the rising number of construction jobs but beneficiaries moving into work were also contributing, MBIE’s quarterly “job-matching” report said. \\
\hline
\end{tabularx}
}

\caption{Provided examples for the idea sentence ($s_p$): \textbf{``There is a
    shortage of construction workers.''}
    The candidate sentence is in bold, with the preceding and following
    sentences provided for context.
\label{tab:candidateex}}
\end{table*}

\begin{minipage}[t]{\linewidth}
\hspace{-0.025\linewidth}  % hackery
\begin{minipage}{0.45\linewidth}
\begin{table}[H]
{\small
\begin{tabularx}{\linewidth}{|l|X|}
\hline
\textbf{Score} &  \textbf{Guidance}: \emph{The candidate sentence\dots}  \\
\hline \hline
1  & \dots is completely unrelated to the idea sentence. \\
\hline
2  & \dots is tangentially related to the idea sentence. \\
\hline
3  & \dots is related to but does not adequately express the idea sentence. \\
\hline
4  & \dots almost expresses the idea sentence. \\
\hline
5  & \dots expresses the idea sentence in its entirety. \\
\hline
\end{tabularx}
}

\caption{Scoring guidelines for the user studies.
\label{tab:scoring} }
\end{table}

\end{minipage}%
\hspace{0.05\linewidth}%
\begin{minipage}{0.45\linewidth}
\begin{figure}[H]
\input{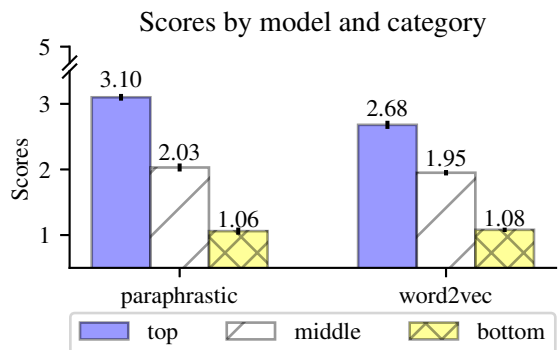}
\caption{User study scores for output from the word-vector-averaging models (\S\ref{subsec:simple-expert}). Error bars represent standard error.
    \label{fig:user-res-wv}}
\end{figure}
\end{minipage}
\end{minipage}

\paragraph{Results.}
Our findings, shown in Fig.~\ref{fig:user-res-wv}, confirmed our expectations:  
users rated \emph{bottom} sentences low (around 1), and \emph{top} sentences better than \emph{middle} ones.
As in \S\ref{subsec:simple-existing}, the paraphrastic vectors led to output receiving better ratings than word2vec (3.1 vs.~2.7 on average), establishing a baseline that finds sentences ``related to, but not (yet) adequately expressing'' $s_p$.

\section{Entailment Models \& Experiments}
\label{sec:complex}

In \S\ref{sec:simple}, we found that word-vector-averaging models
perform only adequately in the disaster recovery application we introduced.
We next consider a model based on a richer notion of semantic matching,
where a matched sentence should \emph{entail} the proposition query.

\subsection{Tree Edit Models}
\label{subsec:complex-model}

As a starting point for the semantic matching function $m(s_p, s)$, we use the tree edit model introduced by \citet{Heilman:10}.  We select this model because it is simple and interpretable, and it was demonstrated to be suitable for a range of semantic similarity problems, including entailment, paraphrase, and answer ranking for question answering.

\subsubsection{Base Model}

We summarize the base model from \citet{Heilman:10} and refer the reader to the original paper for further details.

For the sentences $s$ and $s_p$, we first obtain dependency parse trees\footnote{We use the Stanford CoreNLP pipeline \citep{Manning:14} to obtain dependency parses, lemmas, and part of speech tags.} $T$ and $T_p$, respectively.
We then choose a tree edit sequence (i.e., a sequence of edit operations) that transforms $T$ into $T_p$.
Edit operations include adding nodes (words), deleting nodes, relabeling dependency relations, and so on; the full list is provided in the appendix (Table~\ref{tab:treeops}).
The edit sequence is found using beam search, with a heuristic function that depends on the lemmas, part of speech tags, arc labels, and whether a node is a left or right child of its parent.

A set of 33 integer-valued features are extracted from the edit sequence. These features include the sequence length and counts of different edit types;
the full list is provided in the appendix (Table~\ref{tab:treefeatures}).
A logistic regression (LR) model is trained on these features.

\subsubsection{Neural Tree Edit Model}

Given the many successes of non-linear models and the sequential nature of the tree edits, we introduce a neural network variant of the model. We select a tree edit sequence exactly as described above, and then use a LSTM (\citealp{Hochreiter:97}) that estimates $m(s_p, s)$ by reading
in the tree edits in sequence.  Each element in the tree edit sequence is vectorized as the concatenation of:
\begin{itemize}
    \item A one-hot encoding of the operation type.
    \item A word-embedding-like vector, in the same space as the word embeddings, that aims to capture the word-embedding-space ``difference'' between the sentences before and after the edit operation.  For example, if a new node is added to the tree (\texttt{INSERT-CHILD}, \texttt{INSERT-PARENT}), then we use the word embedding for that word.  If a node is relabeled (\texttt{RELABEL-NODE}) with a new lemma, then we use the difference between word embeddings for the replacement and original word. If a word is deleted (\texttt{DELETE-LEAF}, \texttt{DELETE-\&-MERGE}), then we use the negated embedding of the deleted word.  In other cases, we use a zero vector.
\end{itemize}

This approach allows the model to take lexical and sequential information into account rather than just counts of operations.  Note that both approaches make use of syntactic context when representing edits to sentences.

\subsubsection{Training}

We use the Stanford Natural Language Inference corpus (SNLI; \citealp{Bowman:15}).\footnote{This study had begun before the multi-domain version of the SNLI corpus, MultiNLI \citep{Williams:18}, was released; however, based on post-hoc experiments in \S\ref{subsec:complex-other}, we suspect this would not have made significant impact. Preliminary testing showed that paraphrase corpora (like the MSR Paraphrase Corpus; \citealp{Dolan:04}) were a poor fit.}
SNLI contains approximately 570,000 pairs of sentences (premise and hypothesis); each sentence pair is human-annotated with an \emph{entailment}, \emph{contradiction}, or \emph{neutral} label of the relationship between the two sentences. (As is standard, we ignore examples marked as ``unlabeled'' due to annotator disagreement.)

For the purposes of our matching function $m$, we recast the SNLI examples into a binary framework as follows.
We treat the premise sentence as analogous to the candidate $s$ and the hypothesis as the proposition query $s_p$.
Premise-hypothesis pairs labeled as entailment are considered positive matches, and those labeled as contradiction or neutral are considered negative matches.

We train three model variants: the original logistic regression (LR) model, and the LSTM using the two pre-trained word embeddings discussed and motivated in \S\ref{subsec:simple-model}.   We use the standard SNLI train/development splits to tune hyperparameters; for the LSTM models, we optimize using Adam \citep{Kingma:15}.\footnote{While performance on SNLI specifically is not the goal here, our models perform respectably well on the three-way task (best accuracy is 84.7\%).}

\subsubsection{Fast filter}  Many entailment models, including the ones described in this section, require fairly sophisticated semantic analysis, and therefore significant computational expense. Furthermore, many sentences in $C$ can be easily determined not to match $s_p$.
Therefore, we incorporate the word-vector-based matching functions from \S\ref{subsec:simple-model} as a initial fast filtering step on $C$.%
\footnote{Although not directly relevant to the applications that are the focus of this paper, we note that SNLI sentence pairs (regardless of label) tend to obtain higher scores from the word-vector-based models than sentence pairs from the studies. The high similarity within SNLI and MultiNLI sentence pairs is also supported by \citet{Gururangan:18}.  We take this as encouraging evidence for performing this filtering step before applying SNLI-based models.}

Our procedure, then, is to first score every $s \in C$ according to the fast filter,
then take the top $k$ candidates for selection by $m(s_p, s)$.
The full procedure is outlined in Algorithm~\ref{fig:highlevel}.

\begin{minipage}[t]{\linewidth}
\hspace{-0.025\linewidth}  % hackery
\begin{minipage}{0.45\linewidth}
\begin{algorithm}[H]
\caption{Incorporation of word-vector-based matching as a fast filter.
    Hyperparameters include filter width $k$ and output size $n$.  \label{fig:highlevel}}

\textbf{Input:} corpus $C$, proposition query $s_p$

\begin{algorithmic}
\ForAll{$s \in C$}
	\State obtain ``fast'' score  $wv(s_p, s)$
\EndFor
\State take the top $k$ scoring sentences to be $C_{wv}$
\ForAll{$s \in C_{wv}$}
	\State obtain entailment-based score $m(s_p, s)$
\EndFor
\State return the top $n$ scoring sentences ($C_m$)
\end{algorithmic}
\end{algorithm}

\end{minipage}%
\hspace{0.05\linewidth}%
\begin{minipage}{0.45\linewidth}
\begin{figure}[H]
\input{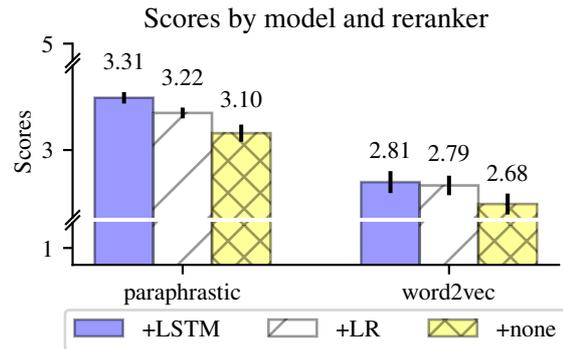}
\caption{User study scores comparing top-25 sentences from just the word-vector-based averaging models (+none) and with reranking from the LR and LSTM tree edit models. Error bars represent standard error.
    \label{fig:user-res-entail}}
\end{figure}
\end{minipage}
\end{minipage}

\subsection{Entailment Model Evaluation}
\label{subsec:complex-eval}

We want to evaluate two hypotheses: (1) that adding the tree edit models on
top of the word-vector-based ones (as described above) yields
better matches; and (2) that using the LSTM-based tree edit model provides
improved performance over the LR-based model.

Our preliminary investigation found that the tree edit models offered
no consistent benefit on the existing-corpora tasks (\S\ref{subsec:simple-existing}).
This is unsurprising; the semantic relationships in those tasks are much broader than 
entailment.
Here we focus entirely on the new, more realistic application in \S\ref{subsec:simple-expert}.
We take the 250 top-scoring sentences from both word-vector-averaging models as ``fast filter'' output, and rerank them using the LR and LSTM tree edit models.
As part of the user study in \S\ref{subsec:simple-expert}, we had users judge the top 25 sentences from the tree edit models' reranked output.

\subsection{Results}
\label{subsec:complex-result}

First, we find that the tree edit model offers some benefit to sentence quality
compared to using only the word vector filters (i.e., the averaging models).
This difference is significant with the paraphrastic filter but within the
range of statistical chance with the word2vec-based filter.

We also find that the LSTM on tree edit sequences offers slightly better
matches than logistic regression; again, this difference is significant with
the paraphrastic-based filter but not the word2vec one.
(In fact, the output from the word2vec filter with
LR and LSTM tree edit models overlaps at about 85\%.)

\paragraph{User feedback.} To gauge interest in the utility of semantic matching
systems, we also asked each user to answer an optional set of
questions after providing judgements. (All users answered the questions.)
We found that (i) 85\% were interested
in a way to measure ideas in news or other corpora, and (ii) half of the
respondents were interested in a follow-up study evaluating semantic matches
from idea sentences of their own choosing.

\subsection{Follow-Up Study}
\label{subsec:follow-up}

Our follow-up study was executed similarly to the the original one described above,
but with proposition queries solicited from users themselves. Instead
of randomly distributing sentences among the follow-up study participants, we
gave each user who participated in the follow-up the output for their own proposition queries.
There were 18 idea sentences and seven participants
in this study. (The full list of idea sentences is provided in Table~\ref{tab:splist2}
in the appendix.) Each participant scored approximately 250
sentences, which were drawn from different parts of the output (as in the
original study).

\paragraph{Results.}
We find that the follow-up study replicates the findings of the original study.
The average scores for the top-ranked output (by the word-vector-averaging models, and reranked by the LR/LSTM models) are generally 0.1-0.2 lower than those in the original study.
However, this decrease holds across different model variants, so the relative performance benefits of using paraphrastic word vectors in the averaging model, as well as using the tree edit LSTM model to rerank, still hold.
We suspect that the decreased scores are partially a function of some of our users' queries being less applicable to the NZ earthquakes (resulting in fewer possible matches), as the emergency managers' expertise is not centered around that particular disaster.

\subsection{Other Entailment Models}
\label{subsec:complex-other}

Because of the limited availability of expert users, we were unable to include a wider range of entailment models in the user study.  It is natural to ask whether alternatives to the model in \S\ref{subsec:complex-model} would have led to better results.
We perform a post-hoc evaluation using the candidate sentences scored by our study participants.
We consider two recent high-performing models: the decomposable attention model (DAM; \citealt{Parikh:16}) and the enhanced sequential inference model (ESIM; \citealt{Chen:17:2}).

To compare performance of these models in this domain, we take all candidate sentences from both the original and follow-up studies (paired with their proposition query) and mark them as ``entailment'' if users scored them with greater than or equal to a 4.\footnote{When a candidate sentence was scored by multiple users, we average their scores.} We split off a set of query-candidate sentence pairs to be a development set; we use these to tune the above models during training (rather than the development sets of SNLI or MultiNLI).

We train these in the two-class setting (entailment vs.~contradiction/neutral) on SNLI; we use existing public implementations for DAM and ESIM.\footnote{DAM: \url{github.com/allenai/allennlp}; ESIM: \url{github.com/nyu-mll/multiNLI}}
We also train these and the LSTM version of the tree-edit model on MultiNLI \citep{Williams:18}, the more recent multi-domain version of SNLI.

\paragraph{Results.}
Table~\ref{tab:other-results} summarizes the $F_1$ scores.
The relatively low performance from all models, despite high performance on SNLI,\footnote{The SNLI website lists DAM and ESIM as having 85\%+ three-way accuracy:\\\url{nlp.stanford.edu/projects/snli/}} indicates that this application is indeed challenging.
We also find that training on MultiNLI instead of SNLI does not offer consistent improvement; that is, the multi-domain nature of that dataset does not seem to improve generalization to our data.
This suggests that our application requires more than modeling sentential entailment.

\begin{minipage}[t]{\textwidth}
\hspace{-0.025\linewidth}  % hackery
\begin{minipage}{0.45\textwidth}
\begin{table}[H]
\begin{tabularx}{\linewidth}{|l|l|X|}
\hline
\textbf{Model} & \textbf{Training data} & \textbf{\emph{F}$_{\mathbf{1}}$} \\
\hline
Tree edit (LSTM/ & SNLI & 55.6 \\
  paraphrastic)  & MultiNLI & 51.3 \\
\hline
DAM & SNLI & \textbf{56.5} \\
    & MultiNLI & 55.2 \\
\hline
ESIM & SNLI & 54.9 \\
     & MultiNLI & 56.0 \\
\hline
\end{tabularx}

\caption{Post-hoc evaluation results (\S\ref{subsec:complex-other}).}
\label{tab:other-results}
\end{table}

\end{minipage}%
\hspace{0.05\textwidth}%
\begin{minipage}{0.45\textwidth}
\begin{figure}[H]

\input{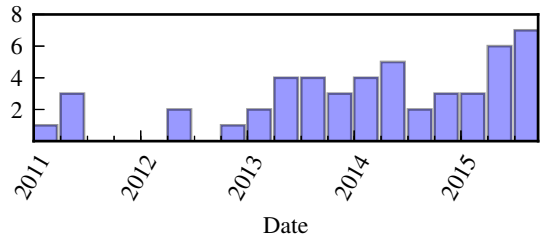}

\caption{Example of semantic measurement: frequency (3 month intervals) of the idea \textbf{``Dealing with authorities is causing stress and anxiety.''}}
\label{fig:measurementex}
\end{figure}

\end{minipage}
\end{minipage}

\section{Semantic Measurement}
\label{sec:measurement}
% figure placed in the same row as the model results table
% \input{histogram-fig}

In this section, we propose an application of obtaining semantic matches of ideas: measuring the frequency of an idea in a corpus across an independent variable (e.g., time).%
\footnote{For more measurement examples drawn from our earthquake recovery data, please refer to \citet{Lin:18}.}
To demonstrate this, we return to the example query from Fig.~\ref{fig:introex}: ``Dealing with authorities is causing stress and anxiety.''
We select this example because it is not easily expressed through $n$-grams, and its output was one of the most highly scored in our user study.
We take the top 50 matched sentences from the paraphrastic vector + tree edit (LSTM) system, determine the publication dates of their source articles via metadata, and compute frequencies in bins of three months.

Our system detects an upward trend in expressions of this idea.  To our domain
expert, this is an interesting yet explainable finding:
in the short term after the earthquake, the focus is more on immediate response and relief.
It takes time for frustration to set in among the population (e.g., due to dealing with bureaucracy and denied insurance claims).
Furthermore, as recovery efforts stretch across years, the media may be more inclined to bring individual stories of continued distress to the forefront.
Future work on semantic measurement could include tuning of hyperparameters (filter width $k$ and output size $n$) and measurement calibration.

\section{Discussion}
\label{sec:discussion}
We discuss some findings from our applications of semantic matching and potential future work.

\paragraph{Desired matches.}
The granularity of the desired matches varies between the applications we presented.
For example, in the framing case, codebook examples are often phrased very generally (e.g., ``supporting immigrants is the moral thing to do''), and evocations of this idea may diverge too much to be detectable by current semantic matching models.
As a consequence, we found that for the two tasks based on existing datasets (\S\ref{subsec:simple-existing}), the entailment-based models from \S\ref{subsec:complex-model} did not help performance (and sometimes hurt).

In contrast, the disaster recovery application demands more specific semantic matches; users were less sure about scoring sentences where the idea was only partially expressed. From both our user study and post-hoc evaluation with other models, we found that while entailment models offer small improvements over the word-vector-averaging baselines, our application requires more than detecting sentence-level entailment.

\paragraph{Entities.}  Particularly in the disaster recovery application, corpus-specific entities can be very important.  Entities like government agencies and insurance companies may be central to queries of interest but lack appropriate distributed representations (sometimes even in the Google News word2vec case) or presence in the training corpus.
(A frequent example in our earthquake news corpus is the Canterbury Earthquake Recovery Authority, often written as ``Cera'' and conflated with the actor Michael Cera.)

\paragraph{Context and coreference.}
Currently, we do not take multiple sentences into account at once when determining sentence matches.
(The user study in \S\ref{subsec:simple-expert} provided context in the survey for the users alone; SNLI deals with this issue by grounding both premise and hypothesis in a specific scenario from an image caption, which is an approach not available in our setting.)
In some cases, this leads to the system finding a match at the sentence level when it would otherwise be invalid from context; in others, a potential match is spread across a sentence boundary.  In future work, including larger and smaller passages (not only sentences) may be worthwhile, especially if coupled with more preprocessing (e.g., coreference resolution, entity linking).

\section{Related Work}
\label{sec:related}
The semantic matching applications in this paper are reminiscent of several lines of research in NLP.

\paragraph{Retrieval.}
As mentioned in \S\ref{sec:problem}, finding coarse semantic matches of a proposition in a corpus is closely related to past work in IR, particularly sentence retrieval \citep{Balasubramanian:07}.
Other relevant work in IR includes \emph{passage} retrieval, which is a component in many web-scale question answering systems \citep{Tellex:03}.
The main difference is that, here, we seek more than a single answer to a question-query; we seek \emph{all matches} to the query (which is a proposition).
Our fast filter also resembles recent work on question answering known as \emph{machine reading} on passages already retrieved \citep{Chen:17}.

\paragraph{Entailment and related tasks.}
There is a long line of entailment tasks and corpora: among others,
the Recognizing Textual Entailment challenges (RTE; beginning with \citealp{Dagan:06});
the Sentences Involving Compositional Knowledge dataset (SICK; \citealp{Marelli:14});
the SNLI and MultiNLI datasets used here (\citealp{Bowman:15}; \citealp{Williams:18});
and the SciTail dataset \citep{Khot:18}.
The RTE-5 through RTE-7 shared tasks, starting with \citet{Bentivogli:09}, contain a similar task to ours; however, these have a very different end goal (using entailment models to improve text summarization) and much smaller corpora (10 documents).

Other related NLP tasks which involve semantic comparisons between pairs of sentences include identifying paraphrase pairs \citep{Dolan:04,Dolan:05} and semantic textual similarity (STS, beginning with \citealp{Agirre:12}).
Both paraphrase and STS differ from entailment (and our semantic matching applications) in that they require bidirectional equivalence; STS furthermore treats similarity on a graded scale rather than as a binary label.

\paragraph{Measurement or tracking of ideas.}
Tracking or measurement of ideas in corpora has often been considered in a more exploratory way, without a user-generated query.  Such exploration has long been a motivation for topic models (e.g., \citealp{Blei:06}).
For example, \citet{Prabhakaran:16} use topics and their rhetorical roles in scientific journal abstracts to understand when topics are in growth or decline.
Other work has allowed user specification of a particular query, though usually as an $n$-gram, as by
\citet{Michel:11}, or using keywords or topics \citep{Tan:17} or short meme phrases \citep{Leskovec:09}.  
We define matches at a more fine-grained proposition level.

\section{Conclusion}
\label{sec:conclusion}
We introduced and explored a new application of semantically matching a proposition against a corpus.  Our findings show that this problem is different from our initial benchmarks based on convenient existing corpora, and from the textual entailment problem.  Our study identified a potential user community and illustrated some factors that will be important in future work.

\section*{Acknowledgements}

This work was supported by NSF \#1541025. LHL was also supported in part by a
NSF Graduate Research Fellowship. Many thanks to the domain experts who
participated in the user studies; Ryan Georgi for the CNN/Daily Mail dataset
suggestion; Dallas Card for help with the Media Frames Corpus; and members of
the ARK and UW NLP for their comments on earlier drafts.

% *****************************************************************
\bibliographystyle{acl_natbib_nourl}
\bibliography{refs}
% *****************************************************************

\pagebreak  \appendix
\section{Appendix}

\subsection{Proposition Queries}

In Table~\ref{tab:framelist}, we provide the thirty proposition queries and associated frames used in the Media Frames Corpus-based evaluation (\S\ref{subsec:simple-existing}).
Table~\ref{tab:splist} lists the twenty proposition queries used in the original user study (\S\ref{subsec:simple-expert}), and Table~\ref{tab:splist2} lists the proposition queries generated by some of our study respondents; these were used in the follow-up study (\S\ref{subsec:follow-up}).

\begin{table*}[!htbp]
{\small
\begin{tabularx}{\linewidth}{|l|X|}

\hline
\textbf{Frame} & \textbf{Proposition query} ($s_p$): \\
\hline \hline
Crime \& punishment & Punishments should be softer on immigration. \\ \hline
Crime \& punishment & Immigrants are no more likely to engage in criminal activity. \\ \hline
Crime \& punishment & Immigrants are more likely to deal or transport drugs. \\ \hline
Cultural identity & Immigrants do assimilate and have similar values to us. \\ \hline
Cultural identity & Immigrants are taking over the country. \\ \hline
Cultural identity & Immigrants have conflicted loyalties and nationalistic sentiments. \\ \hline
Economic & Highly skilled workers are attracted to work in the United States. \\ \hline
Economic & Immigrants work for less money, driving the wages down for domestic workers. \\ \hline
Economic & Immigrants often pay into the system, but do not qualify to receive government benefits. \\ \hline
Fairness \& equality & Immigration rules have changed unfairly over time. \\ \hline
Fairness \& equality & Immigrants cannot wait for the system in place because it is not fair. \\ \hline
Fairness \& equality & Allowing unauthorized immigration is unfair to those who apply and wait. \\ \hline
Fairness \& equality & Law enforcement officials use racial and ethnic stereotypes to unfairly discriminate. \\ \hline
Fairness \& equality & The penalties for illegal immigration should fall on the individuals breaking the law, not businesses. \\ \hline
Health \& safety & Immigrants aid law enforcement by acting as witnesses. \\ \hline
Health \& safety & Immigrants who try to enter the country illegally are responsible for any safety hazards they incur. \\ \hline
Legality \& constitutionality & For free trade to be successful, there should be a free movement of people. \\ \hline
Legality \& constitutionality & Right to work does not mean right to cross national borders. \\ \hline
Legality \& constitutionality & The regulation of immigration should be done through Congress. \\ \hline
Morality & It would be immoral to turn our backs on those in need. \\ \hline
Morality & We have no moral obligation to help those who break the law. \\ \hline
Morality & No path to citizenship creates permanent second-class citizens. \\ \hline
Morality & Supporting the poor does not mean supporting immigrants. \\ \hline
Politics & The immigration issue is a way for politicians to pander to the Hispanic community. \\ \hline
Politics & Businesses have a legitimate interest in lobbying for immigration issues. \\ \hline
Public sentiment & The public supports immigration rights. \\ \hline
Public sentiment & Public support for immigration should not influence policy. \\ \hline
Quality of life & Immigrants drive up the cost of living. \\ \hline
Quality of life & Immigrants have a positive impact on diversity in the United States. \\ \hline
Quality of life & Immigrants deserve better quality of life than they can get in their home countries \\ \hline
\end{tabularx}
}

\caption{Proposition queries used in the Media Frames Corpus evaluation.}
\label{tab:framelist}
\end{table*}

\clearpage

\begin{table*}[!htbp]
\begin{tabularx}{\linewidth}{|X|}
\hline
\textbf{Proposition query} ($s_p$): \\
\hline \hline
Residents are frustrated by the slow pace of recovery. \\ \hline
The repair programme is on schedule to be completed. \\ \hline
Money for repairs is running out. \\ \hline
The council should have consulted residents before making decisions. \\ \hline
Mental health rates have been rising. \\ \hline
Dealing with authorities is causing stress and anxiety. \\ \hline
Most eligible property owners have accepted insurance offers. \\ \hline
Confidence in Cera has been trending downwards. \\ \hline
Water quality declined after the earthquakes. \\ \hline
The power system was fully restored quickly. \\ \hline
Cera missed several recovery milestones. \\ \hline
Prices levelled off as more homes were fixed or rebuilt. \\ \hline
People are suffering because they've lost the intimacy of their relationships. \\ \hline
Coordination between rebuild groups has been problematic. \\ \hline
Few people said insurance companies had done a good job. \\ \hline
Having the art gallery back makes the city feel more whole. \\ \hline
Scirt has spent less money than predicted. \\ \hline
Traffic congestion was severe due to road repairs. \\ \hline
Some of the businesses forced out by the earthquake are returning. \\ \hline
Some of the burden on mental health services is caused by lack of housing. \\ \hline
\end{tabularx}

\caption{Proposition queries used in the original user study (\S\ref{subsec:simple-expert}).
(``Cera'' is short for the ``Canterbury Earthquake Recovery Authority'', and ``Scirt'' is short for the ``Stronger Christchurch Infrastructure Rebuild Team.'')}
\label{tab:splist}
\end{table*}

\begin{table*}[!htbp]
\begin{tabularx}{\linewidth}{|X|}
\hline
\textbf{Proposition query} ($s_p$): \\
\hline
The cost of repairs is over budget. \\ \hline
People are worried that rents will rise. \\ \hline
There are many homeless people in need of shelter. \\ \hline
People called on local corporations to provide additional aid. \\ \hline
The city council could not agree on a plan forward. \\ \hline
Economic inequality grew after the earthquake. \\ \hline
There was a shortage of food and water. \\ \hline
Public transit reroutes and delays caused frustration. \\ \hline
Residents demanded accountability from government agencies. \\ \hline
Small businesses are hit hard by rebuild costs and decreased sales. \\ \hline
Access to electricity continues to be unreliable. \\ \hline
People are struggling to get to their jobs. \\ \hline
People feel less safe in the city. \\ \hline
Hospitals have trouble accommodating all those who need health services. \\ \hline
Residents note a greater sense of community within the neighboorhood. \\ \hline
People feel disconnected to the outside world due to unreliable internet access. \\ \hline
Donations continue to flood in. \\ \hline
The earthquake has exacerbated the housing shortage. \\
\hline
\end{tabularx}

\caption{Proposition queries used in the follow-up study (\S\ref{subsec:follow-up}).}
\label{tab:splist2}
\end{table*}

\clearpage

\subsection{Tree Edit Model}

For reference, we provide the full list of tree edit operations (Table~\ref{tab:treeops}) and features used in the logistic regression version of the model (Table~\ref{tab:treefeatures}) described in \S\ref{subsec:complex-model} and \citet{Heilman:10}.

\begin{table*}[!htbp]
{\small
\begin{tabularx}{\textwidth}{|l|X|X|}
\hline
\textbf{Operation} &  \textbf{Arguments}  & \textbf{Description} \\
\hline \hline
\texttt{INSERT-CHILD}
  & node $n$, new lemma $l$, POS $p$, edge label $e$, side $s \in \{\textit{left}, \textit{right}\}$
  & Insert a node with lemma $l$, POS $p$, and edge label $e$ as the last child (i.e., farthest from parent) on side $s$ of $n$.  \\ \hline
\texttt{INSERT-PARENT}
  &	non-root node $n$, new lemma $l$, new POS $p$, edge label $e$, side $s \in \{\textit{left}, \textit{right}\}$
  & Create a node with lemma $l$, POS $p$, and edge label $e$. Make $n$ a child of the new node on side $s$. Insert the new node as a child of the former parent of $n$ in the same position.  \\ \hline
\texttt{DELETE-LEAF} &  leaf node $n$  & Remove the leaf node $n$.	\\ \hline
\texttt{DELETE-\&-MERGE} &  node $n$ (where $n$ has exactly 1 child)
  & Remove $n$. Insert its child as a child of $n$'s former parent in the same position.  \\ \hline
\texttt{RELABEL-NODE} &  node $n$, new lemma $l$, new POS $p$
  & Set the lemma of $n$ to be $l$ and its POS to be $p$.  \\ \hline
\texttt{RELABEL-EDGE} &  node $n$, new edge label $e$
  & Set the edge label of $n$ to be $e$.  \\ \hline
\texttt{MOVE-SUBTREE}
  & node $n$, node $m$ (s.t. $m$ is not a descendant of $n$), side $s \in \{\textit{left}, \textit{right}\}$
  & Move $n$ to be the last child on the s side of $m$.  \\ \hline
\texttt{NEW-ROOT} &  non-root node $n$, side $s \in \{\textit{left}, \textit{right}\}$
  & Make $n$ the new root node of the tree. Insert the former root as the last child on the s side of $n$.  \\ \hline
\texttt{MOVE-SIBLING}
  & non-root node $n$, side $s \in \{\textit{left}, \textit{right}\}$, position $r \in \{\textit{first}, \textit{last}\}$
  & Move $n$ to be the $r$ child on the s side of its parent.  \\ \hline
\end{tabularx}
}

\caption{Tree edit operations \citep{Heilman:10}.}
\label{tab:treeops}
\end{table*}

\begin{table*}[h]
{\small
\begin{tabularx}{\linewidth}{|l|X|}
\hline
\textbf{Feature category} &  \textbf{Description}  \\ \hline \hline

General counts
  & \# of edits in the sequence; \#s of X edits (where X is one of the operations in Table \ref{tab:treeops}). \\ \hline

\makecell[lt]{\texttt{INSERT-CHILD}, \\ \texttt{INSERT-PARENT}}
  & \#s of these which: insert nouns or verbs, insert proper nouns.  \\ \hline

\makecell[lt]{\texttt{DELETE-LEAF}, \\ \texttt{DELETE-\&-MERGE}}
  & \#s of these which: remove nouns or verbs, remove proper nouns, remove nodes with subject edge labels, remove nodes with object edge labels, remove nodes with verb complement edge labels, remove nodes with root edge labels (which may occur after \texttt{NEW-ROOT} edits). \\ \hline

\texttt{RELABEL-NODE}
  & \#s of these which: preserve POS, preserve lemmas, convert between nouns and pronouns, change proper nouns, change numeric values by more than 5\% (to allow rounding).  \\ \hline

\texttt{RELABEL-EDGE}
  & \#s of these which: change to or from subject edge labels, change to or from object edge labels, change to or from verb complement edge labels, change to or from root edge labels.  \\ \hline

Unedited node counts
  &  In total, numeric values, verbs, nouns, proper nouns.  \\ \hline

Other &  If a tree edit sequence was found or not.  \\
\hline
\end{tabularx}
}

\caption{A description of the tree edit features for LR classification \citep{Heilman:10}.}
\label{tab:treefeatures}
\end{table*}

\end{document}